*Article**Article*

# Classification via an Embedded Approach

José de Jesús Rubio [1,*], Francisco Jacob Avila [2], Adolfo Meléndez [2], Juan Manuel Stein [2], Jesús Alberto Meda [3] and Carlos Aguilar [4]

1. Sección de Estudios de Posgrado e Investigación, ESIME Azcapotzalco, Instituto Politécnico Nacional, Mexico D.F. 02250, Mexico
2. Tecnológico de Estudios Superiores de Ecatepec, Ecatepec, Estado de Mexico 55210, Mexico; favila11@udavinci.edu.mx (F.J.A.); amelendez11@udavinci.edu.mx (A.M.); jstein11@udavinci.edu.mx (J.M.S.)
3. Sección de Estudios de Posgrado e Investigación, ESIME Zacatenco, Instituto Politécnico Nacional, Mexico D.F. 07738, Mexico; jmedac@ipn.mx
4. Centro de Investigación en Computación, Instituto Politécnico Nacional, Mexico D.F. 07738, Mexico; carlosaguilari@cic.ipn.mx
* Correspondence: rubio.josedejesus@gamil.com or jrubioa@ipn.mx; Tel.: +52-55-5729e6000 (ext. 64497)
Received: 5 August 2017; Accepted: 7 September 2017; Published: 15 September 2017**Abstract:** This paper presents the results of an automated volatile organic compound (VOC) classification process implemented by embedding a machine learning algorithm into an Arduino Uno board. An electronic nose prototype is constructed to detect VOCs from three different fruits. The electronic nose is constructed using an array of five tin dioxide ($SnO_2$) gas sensors, an Arduino Uno board used as a data acquisition section, as well as an intelligent classification module by embedding an approach function which receives data signals from the electronic nose. For the intelligent classification module, a training algorithm is also implemented to create the base of a portable, automated, fast-response, and economical electronic nose device. This solution proposes a portable system to identify and classify VOCs without using a personal computer (PC). Results show an acceptable precision for the embedded approach in comparison with the performance of a toolbox used in a PC. This constitutes an embedded solution able to recognize VOCs in a reliable way to create application products for a wide variety of industries, which are able to classify data acquired by an electronic nose, as VOCs. With this proposed and implemented algorithm, a precision of 99% for classification was achieved into the embedded solution.

**Keywords:** Arduino; artificial intelligence; electronic nose; embedded systems; approach; VOC classification
## 1. Introduction

Volatile organic compound (VOC) classification is an important area in a wide range of industries like cosmetics, food, beverages and medical diagnosis, among others [1]. VOC detection could be done through an array of gas sensors conformed as an electronic nose [2] where a data acquisition module converts sensor signals to a standard output to be analyzed and classified. To facilitate VOC detection in situ, an embedded low-power device is required for portable solutions, as well as reliable classification in real time [3].

To decide the right configuration for the electronic nose, an extensive analysis is conducted to determine the number of sensors to conform the array according to the strategies implemented in [4].

Analyzing some methods to build classifiers based on evolution rules as in [5] an electronic nose using a proposed design algorithm as the classifier module into the Arduino is considered to create such a solution. This analysis led to consider a performance comparison among an ATMega 328

*Designs* **2017**, *1*, 7; doi:10.3390/designs1010007 www.mdpi.com/journal/designs

microcontroller device versus a computer running the classifier selected with strategies like in [6] for a portable optimal solution.

There is some research about the classification with intelligent systems. In [1], a learning approach to train uninorm-based hybrid approaches is suggested [7]. Four semi-supervised learning methods are discussed in [2]. In [3], a specific ensemble strategy is developed. A dynamic pattern recognition method is proposed in [4]. In [5,6], the utilization of evolving classifiers for activity recognition is described. Hybrid and ensemble methods in machine learning are focused on in [7]. In [8], a granular framework for evolving fuzzy system modeling is introduced. A novel hybrid active learning strategy is proposed in [9]. In [10], an enhanced version of the evolving participatory learning approach is developed. A class of hybrid-fuzzy models is designed in [11]. A parsimonious approach based on fuzzy inference is addressed in [12]. In [13], a novel dynamic parsimonious fuzzy approach is considered. An evolving hybrid fuzzy based modeling approach is introduced in [14]. A multi-sensor microarray of electronic nose type is created using $SnO_2$ thin film segmented by co-planar electrodes in [8], and the possibility to implement a hardware approach to carry out pattern recognition of signal generated by microarray. In [9], a discussion about the future of the electronic nose turns promising because technicians and researchers around the world are focused on developing innovative instrumental techniques and pattern recognition tools.

In this research, an embedded approach based on an optimal algorithm designed by authors for Arduino board is used to detect and classify a set of VOCs as a portable solution, considering the reduced capabilities of an Arduino instead of a personal computer (PC).

## 2. Approach and Method

To design a method to reach the objective, a problem formulation is made as in [10] and a whole proposal is created for representing the solution. Figure 1 shows the schematic architecture for the portable system to recognize and classify VOCs, where the environment is sensed by the electronic nose conformed by a sensors array. The data acquisition section in the electronic nose receives the sensors array and measured values, which are sent to the approach function to detect patterns and classify those values. This is the electronic nose portable system proposed where the trained approach function is embedded into the Arduino board to accurately and reliably detect and classify a set of VOCs.

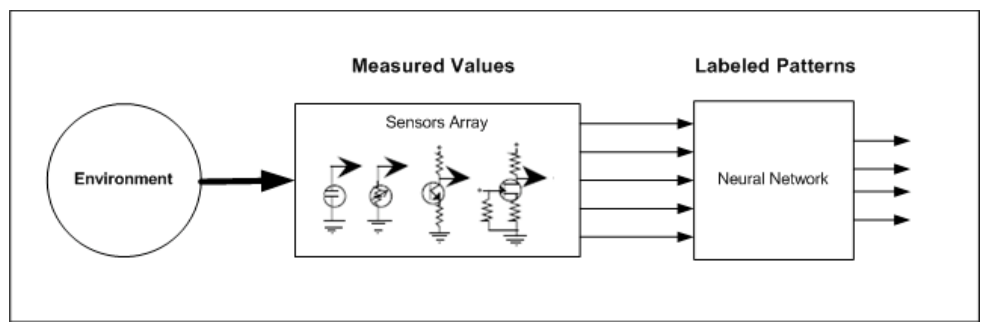

**Figure 1.** Sensor system and a trained approach.

To create the electronic nose, a set of sensor gas is chosen to be able to detect a wide range of volatile organic compounds and get a fast and reliable response at a low cost that is an easily used solution.

Each component element that conforms to the solution is presented in the following subsections, as well as conducted experiments and their results as a portable application with the propose algorithm.

*2.1. Electronic Nose*

An electronic nose is a device designed to detect, identify and quantify chemical vapors as VOCs. To do that, the electronic nose combines gas sensors with a pattern recognition system [11].

The sensors array is created using five semiconductor tin dioxide gas sensors as shown in the Table 1. Where a description of every sensor was collected from its datasheet.

**Table 1.** Gas sensors chosen for the sensors array in the electronic nose.

| Gas Sensors for the Sensors Array | | | |
|---|---|---|---|
| **Model** | **Manufacturer** | **Material** | **Chemical Sensitivity (Collected from Data Sheets)** |
| MQ-2 | Hanwei | $SnO_2$ — Tin dioxide | Combustible gas and smoke, Liquid Petroleum Gas (LPG), isobutane, propane, methane, alcohol, hydrogen |
| MQ-3 | Hanwei | $SnO_2$ — Tin dioxide | Alcohol and Benzene |
| MQ-135 | Hanwei | $SnO_2$ — Tin dioxide | Air quality, $NH_3$ Ammonia, NOx — nitrogen oxide, $CO_2$, alcohol and benzene |
| TGS2610 | Figaro | Semiconductor type D1 | Gas Liquid Petroleum (LP), Propane and Butane |
| TGS2611 | Figaro | Semiconductor type D1 | Methane and Methane |

Every sensor is design for a specific group of chemicals sensitivity, which varies on the load resistor configuration according to the following equation.

$$R_s = \frac{V_C - V_{RL}}{V_{RL}} R_L ,\qquad(1)$$

where *Rs* is the internal sensor resistor and is calculated with the voltage measured in the load resistor $R_L$ in the configuration circuit shown in Figure 2.

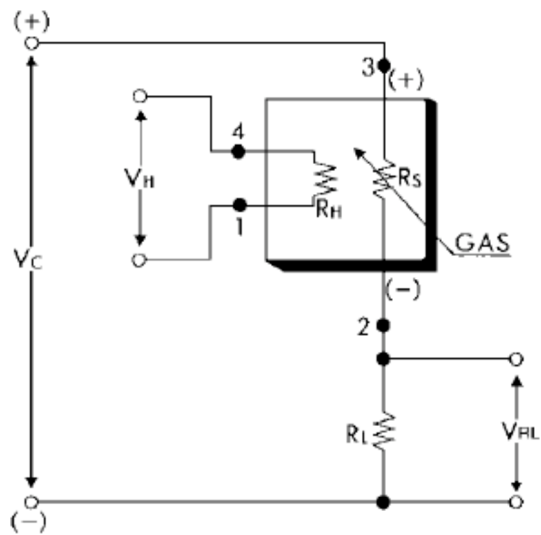

**Figure 2.** Configuration circuit for each gas sensor.

For every gas sensor there is a response time influencing in the whole array response as a combination of the initial response, the steady-ready state response and the dynamic slope of the response, which are often well correlated with the type of pattern [12].

As an electronic nose, the sensor array design is shown in Figure 3.

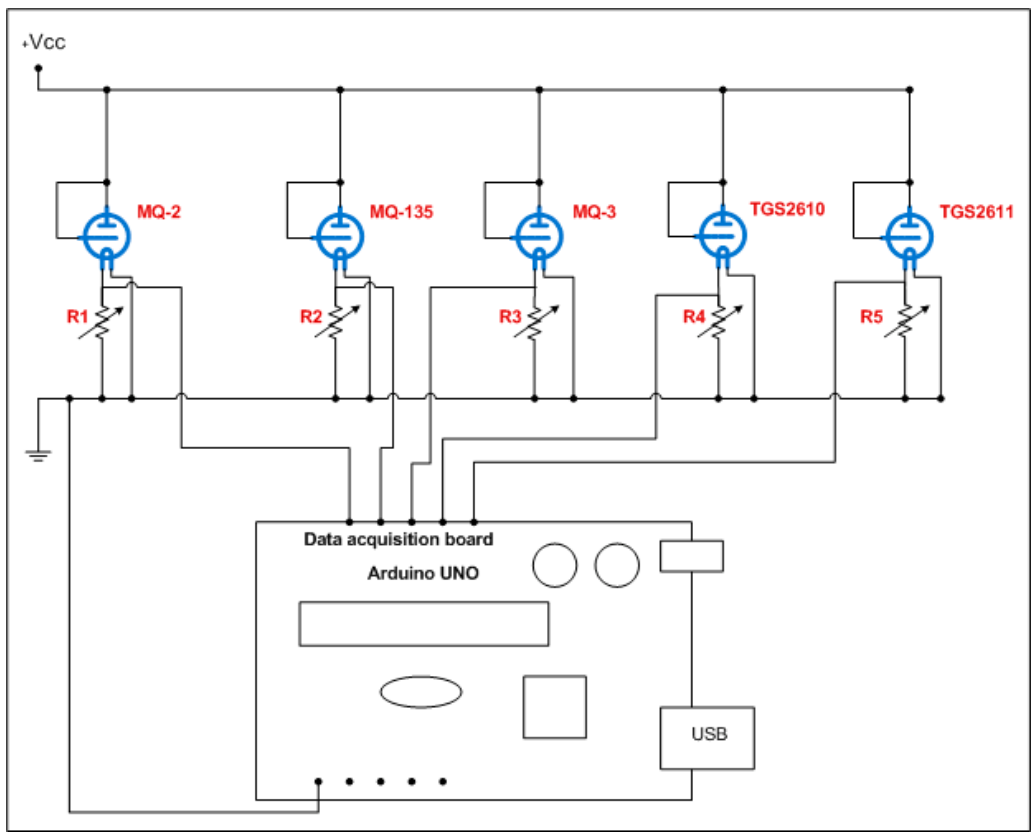

**Figure 3.** Sensor array electronic diagram for the electronic nose.

*2.2. Arduino Uno*

Considering the use of an Arduino Uno as an ATMega 328 microcontroller system instead of a Field Programmable Gate Array (FPGA) as in [13], Arduino is an open electronic platform for easy prototyping and is based on flexible microcontroller hardware and software that allows the creation of intelligent solutions in an easy way [14].

The microcontroller is programmed using the Arduino development environment and the loaded program can be run without a computer. For this project, Arduino Uno board is used for its six analog input pins, which can receive signals from five gas sensors as in [15].

In this project, Arduino Uno will be responsible for receiving all the information of sensors and to acquire sensed values; the Arduino program reads analog inputs every 500 ms and preprocess them to get ready for the classification process in the approach function, then results are shown in a Liquid Crystal Display (LCD) display indicating the identified component. Figure 4 shows the approach which performs the whole process.

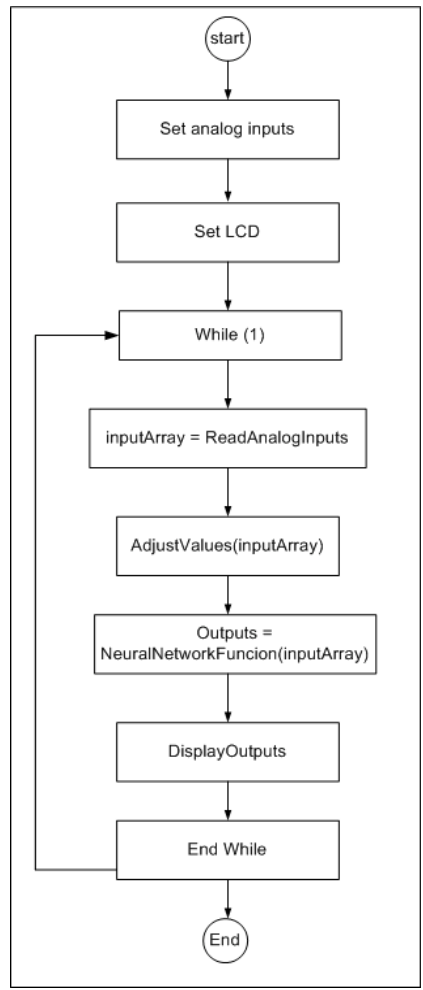

**Figure 4.** Algorithm for the Arduino program to process analog inputs that carry sensor values.

## 2.3. Approach

The approach is a learning paradigm and an automated process based on the animals nervous system. It is composed by a group of interconnected units which collaborate with each other to produce an output [16], Figure 5 shows a general unit with a single input.

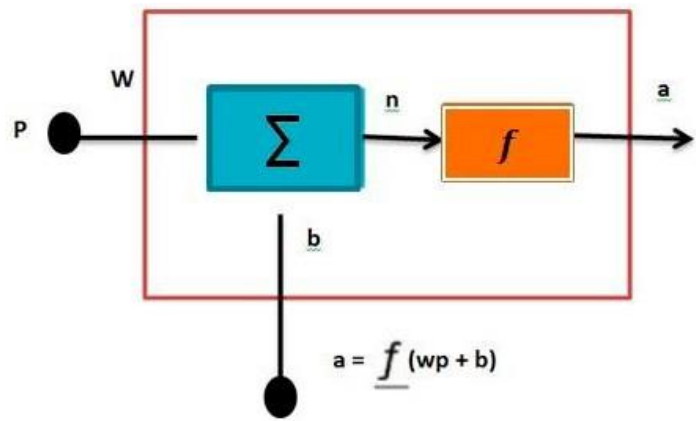

**Figure 5.** A general unit with a single input.

To perform the output *a*, unit process input *p* which is multiplied by the value *w* to create *wp* as the first term into the adder. The second term is *a* value of 1 that is multiplied by *a* bias *b*, which represents an offset. The adder output *n*, which is equal to *wp* + *b*, is known as the net input and is sent to the transfer function, which generates the unit output *a*. In this way:

$$a = f(wp + b), \qquad (2)$$

Variables *w* and *b* are adjustable parameters which are set by the learning rule used to train the approach [17].

In a multi-phase architecture, output units from one phase are connected as inputs for the next phase, as shown in Figure 6.

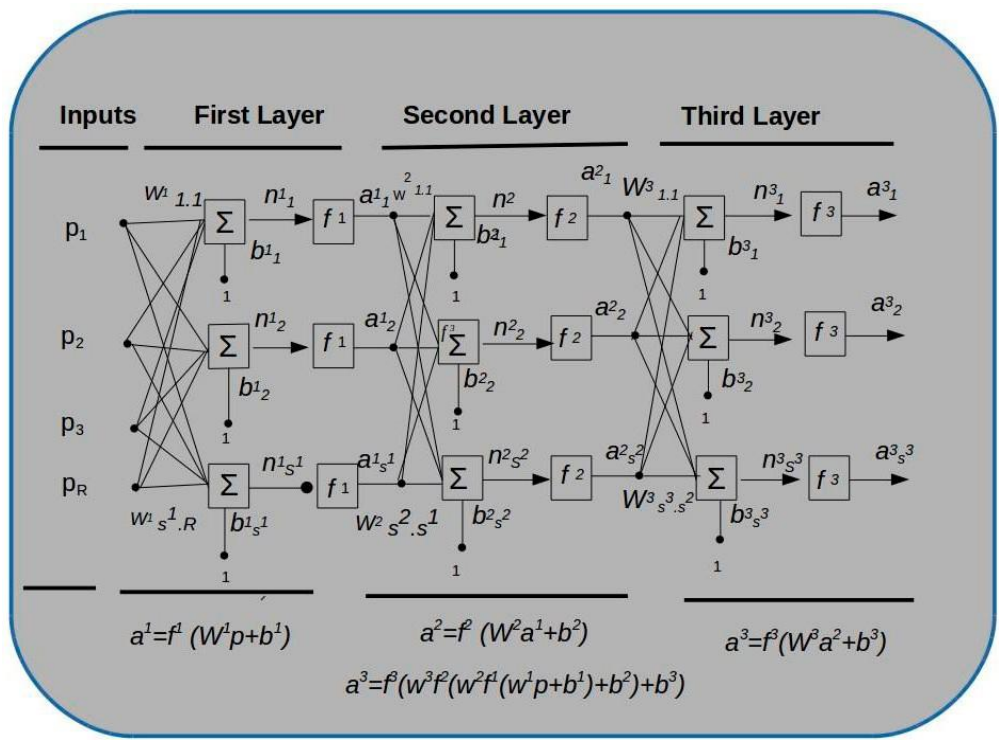

**Figure 6.** Three-phase approach with multiple inputs.

Transfer function would be the same for all units in the same phase, and could be different from every phase. The transfer function is chosen by the designer and, as stated before, *w* and *b* are adjusted by the chosen learning rule to meet specific goals in the relationship between input and output [18].

There are different transfer functions that are chosen for different purposes. As described before, *n* is the net input, which is calculated as *n* = *wp* + *b*, and the unit output is given by *f* (*wp* + *b*) or *f* (*n*) where *f* is the transfer function, which could be a linear or nonlinear function of *n*, and is chosen to satisfy an specification of the problem in which the unit is designed to solve [19].

For the purpose of this project an approach with two phases and a Log-Sigmoid transfer function for both phases is chosen to be implemented into the Arduino. This is because of the reduced capabilities and performance in an Arduino, which are not comparable with a PC in terms of processor and memory. That is why it is better for the Arduino to use the basic configuration with only two phases and Log-Sigmoid as a transfer function, Equation for the Log-Sigmoid transfer function is shown in Figure 7.

| Log-Sigmoid | $a = \dfrac{1}{1 + e^{-n}}$ |
|---|---|

**Figure 7.** Equation for the Log-Sigmoid transfer function.

*2.4. Implementation and Experiments*

The implementation as an electronic nose with the Arduino and a Liquid Crystal Display (LCD) module is constructed as shown in Figure 8, and the approach architecture that is embedded is shown in the Figure 9.

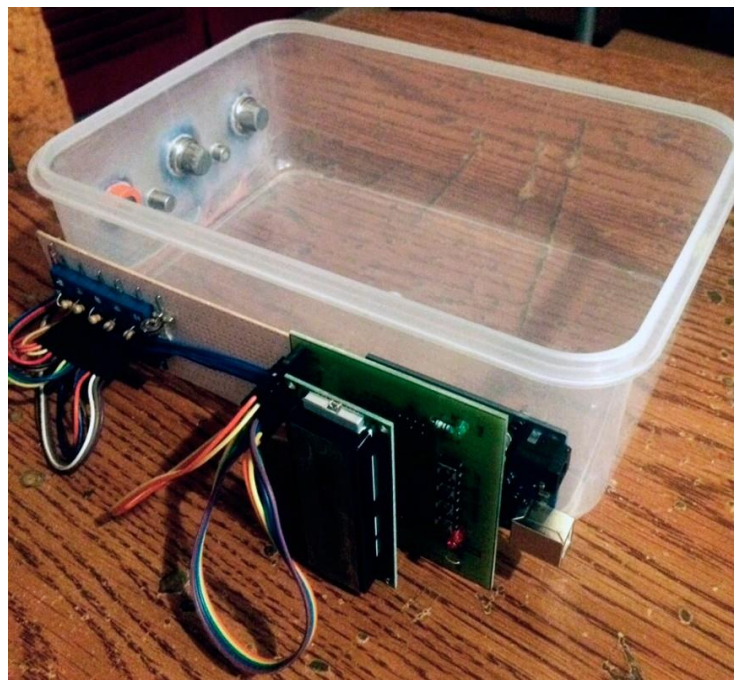

**Figure 8.** Electronic nose implemented with an array of five gas sensors and an Arduino with a Liquid Crystal Display (LCD) board.

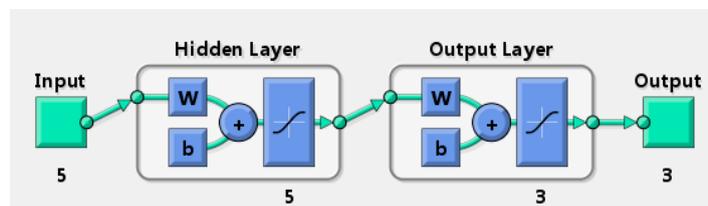

**Figure 9.** Approach architecture created with five inputs, five units in the hidden layer, and three units in the output layer.

The construction of the electronic nose is implemented using a connections board as previously specified in Figure 3, placing gas sensors indicated in the Table 1. The prototype is mounted in a plastic shield as shown in Figure 8. After the construction of the electronic nose, the prototype is first tested and set up to collect data from three different products: lemon, banana and grape.

The data collection is performed following a specific procedure in order to be able to sense three different compounds one at a time without interfering with one another, as described in the following procedure stated in the Table 2.

**Table 2.** Procedure for data acquisition of organic compounds.

| /* Procedure for data acquisition */ |
| --- |
| OrganicCompounds = [Lemon, Banana, Grape] |
| Open TopCover from SensingChamber |
| Turn on electronic nose |
| Wait 10 min |
| For each element in OrganicCompounds |
|     Place element in SensingChamber |
|         Close TopCover of SensingChamber |
|         While time <= 5 min |
|             Capture data from 5 sensors |
|         End while |
|     Open TopCover from SensingChamber |
|     Remove element |
|     Wait 5 min |
| End For |

The data is received in the computer to be stored in a file for the proper adjustment required for inputs and targets needed for the approach training.

Applying the procedure described in Table 2, a total of 2500 sample values were collected in 35 min for the three organic products.

The procedure starts with pre-heating sensors in periods of 10 min; as soon as the first 10 min is concluded, lemon is placed into the sensing chamber and top cover is closed to start the values registration for 5 min, where 689 values are received and registered. Table 3 shows a list of 7 out of 689 values collected in five minutes from 5 sensors for lemon.

**Table 3.** Seven out of 689 values collected in five minutes from five sensors which identify lemon.

| MQ-2 | MQ-135 | TGS2610 | TGS2611 | MQ-3 |
| --- | --- | --- | --- | --- |
| 138 | 64 | 68 | 90 | 111 |
| 139 | 64 | 69 | 90 | 111 |
| 167 | 79 | 93 | 95 | 123 |
| 167 | 77 | 91 | 95 | 124 |
| 168 | 78 | 91 | 96 | 129 |
| 208 | 84 | 112 | 90 | 249 |
| 210 | 85 | 114 | 90 | 252 |

After the 5 min period, the top cover is opened and the lemon is taken off to clean the air into the sensing chamber without the need of an air pump; the top cover is left open for 5 min more to clean the air without the need of purge pump. At this point, just 20 min has passed since the procedure started, and immediately a banana is placed in the sensing chamber, closing the top cover again to collect data for 5 min, and a total of 728 values from banana are received and registered. Concluding this period, the top cover is opened and the banana sample and is taken off, leaving the top cover open for another 5 min for cleaning the air again. Finally, after the last 5 min for cleaning are concluded, the grape is placed in the sensing chamber, closing the top cover to collect data for 5 min to receive 692 values from grape. The procedure concludes 35 min after it started with around 700 samples from each organic compound for a total of 2109 sample data for the three compounds. The difference between each component is due to the elapsed and delay time when the top cover is open and closed, as well as the time for cleaning the sensing chamber.

Once the data is collected from the three different compounds an approach is created in the MATLAB toolbox (R2010a, Mathworks, Natick, MA, USA) with the following characteristics specified in Table 4:

Table 4. Approach characteristics that are constructed in MATLAB toolbox.

| Approach | |
| --- | --- |
| Approach type | Feed Forward |
| Inputs | 5 |
| Phases | 2 |
| Hidden phases units | 5 |
| Transfer function in hidden phase | Log-Sigmoid |
| Output phase units | 3 |
| Transfer function in output phase | Log-Sigmoid |
| Training algorithm | Stochastic gradient descent |

There are five units in the input phase because there are five gas sensors; in this way a five-unit hidden phase is chosen to reduce complexity and because three organic compounds are selected, three outputs are required for the output phase, which means that three units are required in the output phase [20].

To start the training process, the sensors data is arranged as inputs for the approach and outputs as targets are established for each organic compound as follows:

$$p_{Lemon} = \begin{bmatrix} MQ-2 \\ MQ-135 \\ TGS2610 \\ TGS2611 \\ MQ-3 \end{bmatrix}, t_1 = \begin{bmatrix} 1 \\ 0 \\ 0 \end{bmatrix}, \quad (3)$$

$$p_{Banana} = \begin{bmatrix} MQ-2 \\ MQ-135 \\ TGS2610 \\ TGS2611 \\ MQ-3 \end{bmatrix}, t_2 = \begin{bmatrix} 0 \\ 1 \\ 0 \end{bmatrix}, \quad (4)$$

$$p_{Grape} = \begin{bmatrix} MQ-2 \\ MQ-135 \\ TGS2610 \\ TGS2611 \\ MQ-3 \end{bmatrix}, t_3 = \begin{bmatrix} 0 \\ 0 \\ 1 \end{bmatrix}, \quad (5)$$

where $p$ is the input vector of five sensors and $t$ is the target vector for each compound sensed. For lemon the output will be [1, 0, 0], for banana the output will be [0, 1, 0] and for grape its output will be [0, 0, 1].

The results of the training process show an acceptable classification for the elements sensed, performance and confusion matrix indicate a 100% of precision in 107 epochs without false positives.

Figure 10 shows the results of training where the toolbox chose randomly from a complete data sample, 70% of data for training, 15% for testing, and 15% for validation from 2109 total samples from the three compounds. The graphic shows that test, denoted with a red line, under the green and blue ones, which resulted in a lower error, which means that test had a better performance than the training. Figure 11 is showing the confusion matrix where it can be seen that no false positive where generated during training nor testing and validation.

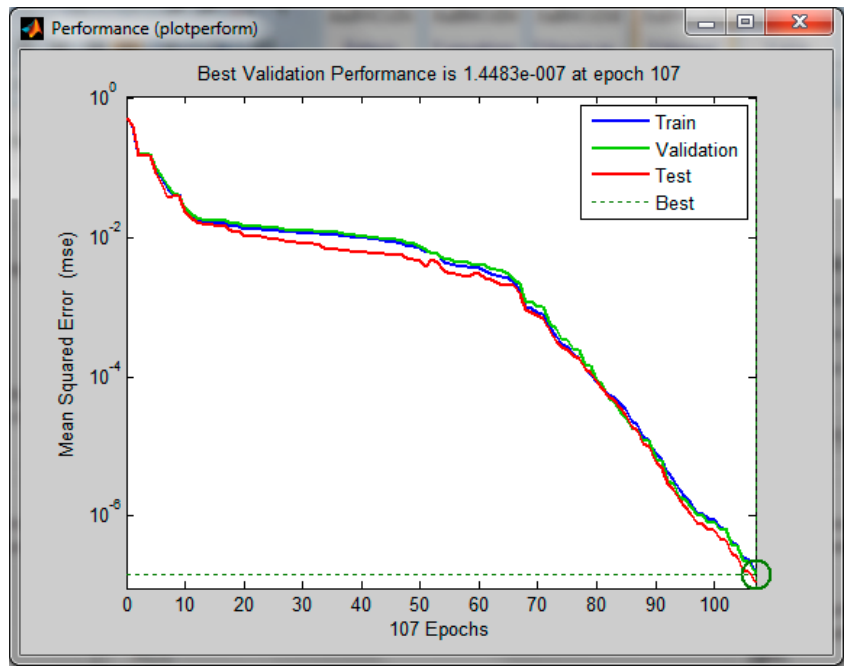

**Figure 10.** Performance for the training process where mean square error of $1.4483 \times 10^{-7}$ is reached at 107 epochs.

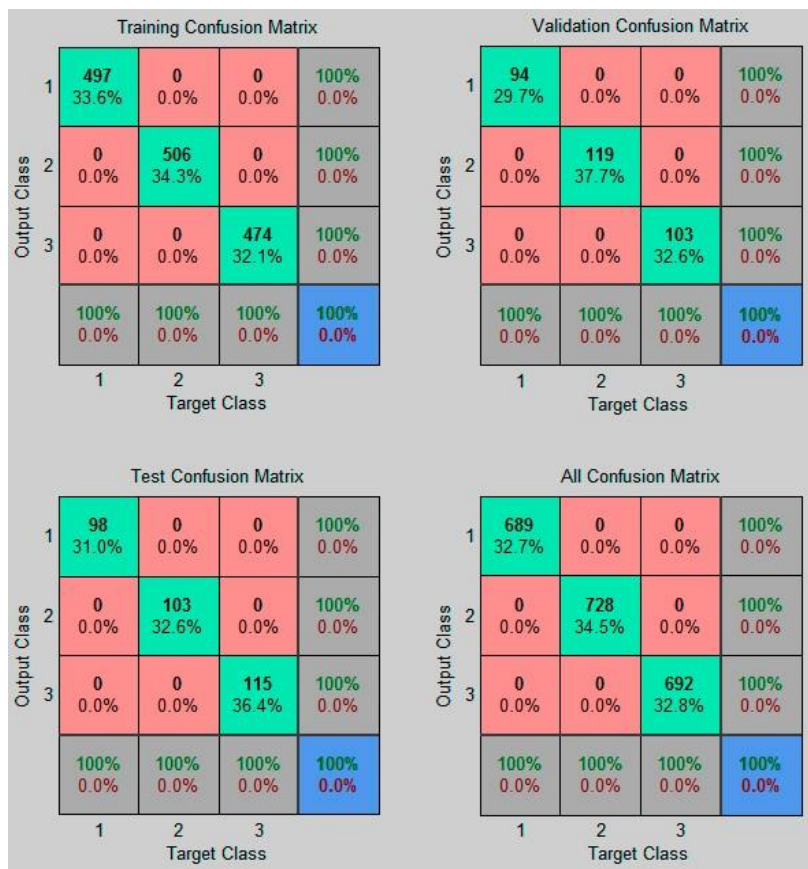

**Figure 11.** Confusion matrix resulted after training process where 100% of data is classified correctly.

This test is the right approach to implement it as an embedded algorithm into the Arduino. To achieve that, the value matrices and bias vectors for hidden and output phases are extracted from the training process and are as follows:

$$HLW = \begin{bmatrix} 0.96053576 & -0.49067116 & -2.25964108 & 2.50108093 & 0.19458625 \\ 0.46907827 & 5.48986523 & -4.78114212 & -4.99721858 & 2.80680594 \\ -1.57636870 & -0.58488740 & 2.68218068 & -1.25514649 & 4.43993330 \\ -0.57512130 & 2.82730697 & 1.04113772 & -5.14422524 & -1.64028299 \\ 2.63760244 & 2.83163383 & 2.89952632 & -0.36016259 & -2.66582192 \end{bmatrix} \quad (6)$$

$$HLB = \begin{bmatrix} -1.02772180 & -1.77963962 & 2.64182372 & -0.97785230 & 1.63223721 \end{bmatrix} \quad (7)$$

$$OLW = \begin{bmatrix} 2.24302499 & -3.92702259 & -2.17715966 & 2.91615488 & -3.41820192 \\ -2.09640007 & 7.71309853 & -4.06959432 & 0.4036855 & 2.39010108 \\ 1.86633665 & -4.39788864 & 6.46096038 & -3.02420647 & -1.04196844 \end{bmatrix} \quad (8)$$

$$OLB = \begin{bmatrix} -2.90288788 & -1.10540564 & 0.49794008 \end{bmatrix} \quad (9)$$

*Where*
*HLW is for Hidden Layer Weights,*
*HLB is for Hidden Layer Bias,*
*OLW is for Outpit Layer Weights and*
*OLB is for Output Layer Bias*

With these values as constants, the approach function implementation to be embedded into Arduino is shown in the approach flow chart of the Figure 12.

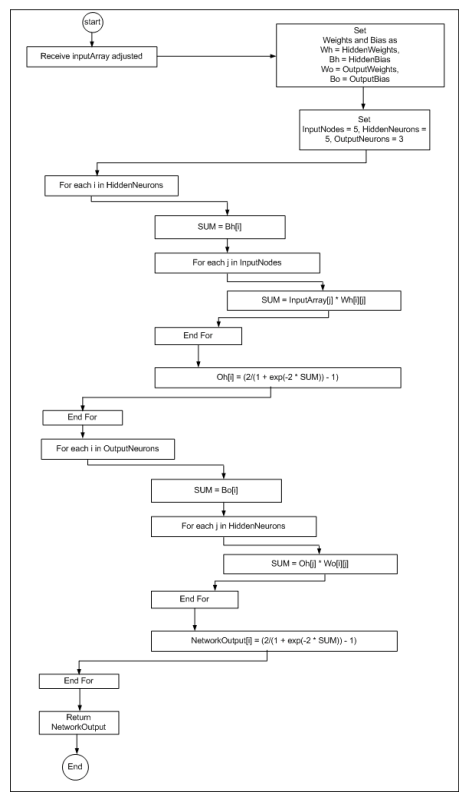

**Figure 12.** Approach embedded in Arduino.

*2.5. Advised Approach*

The revision of multiple approaches [13,21–23] are considered to establish an innovative method and solution for this research, a modification of approach needs to take place to conform this proposal and solution. The approach is some kind of generalization of the Learning Management System (LMS) approach using the same performance index, which is represented by the mean square error, calculated by comparing the approach output with the target output. The approach should adjust the parameters to minimize the mean square error.

$$F(X) = E[e^2] = E[(t-a)^2], \tag{10}$$

where $X$ is the vector of the approach value and bias and when the approach has multiple outputs, this can be generalized to:

$$F(X) = E[e^T e] = E[(t-a)^T(t-a)], \tag{11}$$

Then as with the LMS approach, the mean square error can be approximated by:

$$F(X) = (t(k) - a(k))^T (t(k) - a(k)) = e^T(k)e(k), \tag{12}$$

where the squared error expectation has been replaced by the squared error at iteration $k$.

The stochastic gradient descent is:

$$w_{i,j}^m(k+1) = w_{i,j}^m(k) - \alpha \frac{\partial F}{\partial w_{i,j}^m}, \tag{13}$$

$$b_i^m(k+1) = b_i^m(k) - \alpha \frac{\partial F}{\partial b_i^m}, \tag{14}$$

where $\alpha$ is the learning rate for $w$ and $b$ represents the value and bias, $m$ represents the corresponding phases in the approach, $i$ the input and $j$ the unit in the phase $m$.

Now, the computation of the partial derivatives is a heavy load process for training and could be beyond Arduino's capability. That is where the algorithm modification takes place for an optimum performance to achieve a successful classification in the Arduino.

Because the error is an indirect function of the value in the hidden phase, the chain rule of calculus is going to be used to calculate the derivatives. When there is an explicit function $f$ only of the variable $n$, it is needed to take the derivative of $f$ with respect to a third variable $w$. The chain rule is then:

$$\frac{df(n(w))}{dw} = \frac{df(n)}{dn} \times \frac{dn(w)}{dw}, \tag{15}$$

This concept is employed to find the derivatives in Equations (13) and (14):

$$\frac{dF}{dw_{i,j}^m} = \frac{dF}{dn_i^m} \times \frac{dn_i^m}{dw} \tag{16}$$

$$\frac{dF}{db_i^m} = \frac{dF}{dn_i^m} \times \frac{dn_i^m}{db_i^m} \tag{17}$$

The second term in each of these equations can be easily computed since the input to the phase $m$ is an explicit function of the values and bias in that phase [17].

To implement this, firstly, and using Tanimoto concepts in [24], initial values are considered to be randomly equal to inputs as stated in the following algorithm shown in Figure 13:

**Algorithm 1.** Modified stochastic gradient descent trainng.

TRAIN(inputs, targets)
    SET randombly $W \in$ inputs
    $N_t = |i: t_i = t|\ for\ t = -1, +1$
    $\alpha = \sqrt{MAX(N_{-1}, N_{+1})/N_{t_i}}\ for\ i = 1, \ldots N$
    $s^m = -2F^m(n^m)(t - input)$
    $W^m(k+i) = w^m(k) - \alpha s^m(input^{m-1})^T$
    $bias^m(k+i) = bias^m(k) - \alpha s^m$

    return W, bias

**Figure 13.** Modified stochastic gradient descent training algorithm.

The modification of the algorithm resulted in a slight change for the Arduino; moreover, the initial values are randomly equal to inputs. Values and bias are updated using the stochastic gradient descent rule and sensitivity s is propagated backward through the approach.

The solution is mathematically equivalent with the LMS approach, but with the slight modification presents some significant practical consequences.

## 3. Discussion

The same organic compounds are used to examine the capability of the prototype with the embedded approach and its modified training approach. Image in the Figure 14 shows the lemon identification with the approach output values and, as in [25], a comparative analysis was performed.

The implemented approach works as efficiently as the approach constructed in MATLAB using the same inputs and targets values for training process, which demonstrates an autonomous solution. Changing compounds requires the electronic nose to be left without any element for a while to let VOCs and gas concentrations disperse, but when a new element is placed in the sensing chamber and the top cover is closed, there is an immediate response identifying the compound and showing the result in the LCD display.

The established hidden units responses allow the three types of VOCs to be classified accurately and reliably.

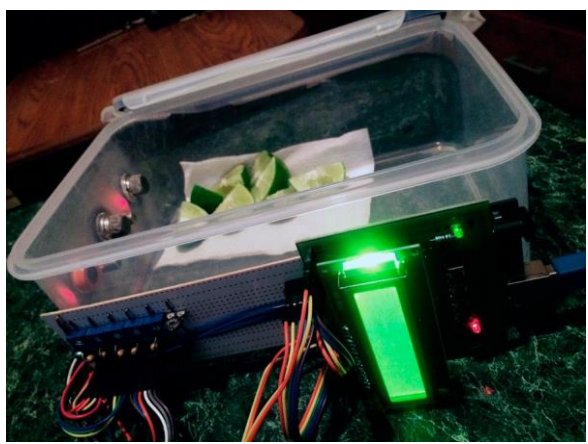

**Figure 14.** Electronic nose prototype sensing lemon and displaying classification by the embedded approach.

As in [26] the sensation of smell resulting from a specific molecular recognition can be used as an analytical tool in many industries to measure, for example the quality of food, drinks or chemical products. The electronic nose developed for this research had the main objective to improve the response to detect VOCs [27] as well as deliver a portable tool.

The purpose of this works satisfied the need for rapid, sensitive and highly portable identification and recognition of VOCs, as well as in [28], the solution shows the discriminatory capacity of sensor array using a patterns recognition approach.

A complex mixture of volatile compounds that exhibit qualities, intensity and different concentrations, due to the complexity of the organic matter can be attended by this kind of innovative technology [29] which seems promising for the future of electronic noses because researchers throughout the world are increasing their perspective and efforts to develop new innovative tools. All of this including embedded algorithms like the proposed approach [30], which are widely implemented to automate solutions.

## 4. Conclusions

An autonomous electronic nose has been designed, constructed and tested. The experimental results show that an embedded approach in an Arduino can be reliable and grant enough precision to classify VOCs, which can be scalable to implement biomedical applications as well as environment monitoring systems. In the future, other kinds of embedded approaches will be designed.

**Acknowledgments:** Authors thank the Instituto Politecnico Nacional, and Tecnologico de Estudios Superiores de Ecatepec for the support in this approach.

**Author Contributions:** José de Jesús Rubio, Jesús Alberto Meda, and Carlos Aguilar conceived and designed the experiments; Francisco Jacob Avila, Adolfo Meléndez, and Juan Manuel Stein performed the experiments; Jesús Alberto Meda, Carlos Aguilar, Adolfo Meléndez, and Juan Manuel Stein analyzed the data; José de Jesús Rubio and Francisco Jacob Avila wrote the paper.

**Conflicts of Interest:** The authors declare no conflict of interest.